\documentclass[10pt]{article} 
\usepackage[preprint]{tmlr}

\makeatletter
\if@accepted
    \newcommand{\codelink}{https://github.com/MikkelGodsk/SAE-labelling}
\else
    \newcommand{\codelink}{https://anonymous.4open.science/r/Sparse-Autoencoders-392b}
\fi
\makeatother

\usepackage{amsmath,amsfonts,bm}

\def\eqref#1{equation~\ref{#1}}

\def\1{\bm{1}}

\DeclareMathAlphabet{\mathsfit}{\encodingdefault}{\sfdefault}{m}{sl}
\SetMathAlphabet{\mathsfit}{bold}{\encodingdefault}{\sfdefault}{bx}{n}
\newcommand{\tens}[1]{\bm{\mathsfit{#1}}}

\usepackage{graphicx} 
\usepackage{url}
\usepackage{hyperref}
\usepackage{amsmath, amssymb, amsfonts}
\usepackage{algorithm, algpseudocode}
\algrenewcommand\algorithmicrequire{\textbf{Input:}}
\algrenewcommand\algorithmicensure{\textbf{Output:}}
\usepackage{rotating}
\usepackage{caption}
\usepackage{subcaption}

\usepackage{booktabs}
\usepackage{multirow}
\usepackage{makecell}
\usepackage{array}
\usepackage{tcolorbox}
\usepackage{float}

\usepackage{listings}
\title{Steering LLMs? Actually, Sparse Autoencoders \textit{can} outperform simple baselines}

\author{\name Mikkel Godsk Jørgensen \email mikkel.godsk.research@gmail.com \\
      \addr DTU Compute\\
      Technical University of Denmark
      \AND
      \name Lars Kai Hansen \email lkai@dtu.dk \\
      \addr DTU Compute\\
      Technical University of Denmark}

\def\month{??}  
\def\year{2026} 
\def\openreview{\url{https://openreview.net/forum?id=XXXX}} 

\begin{document}

\maketitle

\begin{abstract}
    \noindent Sparse Autoencoders (SAEs) have been seen as a promising avenue for exploring the internals of Large Language Models (LLMs) and for steering model output generation. 
    When AxBench - a model steering benchmark - was introduced in \cite{wu2025axbench}, SAEs did not seem to live up to their original hype due to poor steering performance relative to a set of simple baselines. 
    This work serves as a partial rebuttal for Sparse Autoencoders and suggests that the results of \cite{wu2025axbench} did not do them full justice.
    We find that Sparse Autoencoders can, in fact, perform close to on par with the reference LoRA performance on the AxBench benchmark, when features are selected and labelled with our supervised pipeline.
    We also find that our pipeline selects features that are surprisingly causal of their identified labels when using only its interpretability-based components.
    Lastly, we present evidence that high sparsity (low $\ell_0$) may not be crucial for successful steering based on interpretability, which is in contrast to the earlier findings in \cite{wang2025doeshigherinterpretabilityimply}. Our code is available at \url{\codelink}.
\end{abstract}
\section{Introduction} 
Sparse Autoencoders have recently attracted a wave of attention \citep[e.g.][]{cunningham2023sparse,makelov2024towards,templeton2024scaling,gao2024scaling,rajamanoharan2024jumping}, offering an unsupervised interpretability model based on the \textit{Linear Representation Hypothesis} (LRH) and the \textit{superposition hypothesis}. The LRH suggests that many concepts may be stored in linear subspaces, as discussed in e.g. \citet{alain2018understandingintermediatelayersusing,kim2018interpretabilityfeatureattributionquantitative,elhage2022superposition,park2023linear}, although somewhat contested in e.g. \citet{crabbe2022concept}. 
Since LLMs can represent a large number of concepts in relatively low-dimensional latent spaces, the neurons must be \textit{polysemantic} as a consequence, i.e. they become active across multiple different semantics, as discussed in e.g. \citet{bolukbasi2021interpretability,makelov2024towards, templeton2024scaling, fereidouni2025evaluatingsparseautoencodersmonosemantic, cunningham2023sparse}.
The superposition hypothesis further suggests that linear concept representations are densely packed and not fully orthogonal \citep{elhage2022superposition}.
Within this framework, Sparse Autoencoders offer an appealing interpretability tool, capable of discovering a subset of linear concept representations, referred to as \textit{features}, which are thought to exhibit a high degree of \textit{monosemanticity} \citep[e.g.][]{bricken2023monosemanticity,templeton2024scaling}.
In \cite{fereidouni2025evaluatingsparseautoencodersmonosemantic}, the authors quantify the monosemantic property of SAEs in comparison to individual neurons in a network. Their findings suggest that, indeed, SAE features are more monosemantic than the individual neurons of language models.

While the monosemantic response is interesting for interpretability, features have also shown some ability to control the model's text generation
\citep[e.g.][]{templeton2024scaling, arad2025saesgoodsteering, he2025saif, wang2025doeshigherinterpretabilityimply}, although the effectiveness is disputed most notably in \cite{wu2025axbench}. In fact, \cite{wu2025axbench} concludes that ``Even simple baselines outperform Sparse Autoencoders'' for steering, and finds that simply prompting the model is much more effective than other approaches, such as intervention via SAEs \citep[e.g.][]{templeton2024scaling} and LoRA \citep{hu2022lora}. Although this is undeniably important for most practical applications, the prompt baseline can seem unfair since the subject LLM was originally trained precisely to follow instructions.
Aside from steering, there have also been several attempts at using SAEs for knowledge unlearning/concept removal in language models \citep[e.g.][]{farrell2024applyingsparseautoencodersunlearn, yamashita2025sparseautoencoderguidedinternalrepresentationunlearning, fereidouni2025evaluatingsparseautoencodersmonosemantic}, essentially exploring the negative (removal) dimension of steering.

Since Sparse Autoencoders are trained unsupervised, efforts have been made to label their features with the concepts they represent. This has been attempted at scale using LLMs \citep[e.g.][]{lieberum2024gemma,paulo2024automatically}, but to our knowledge, no dedicated efforts have been made to use a supervised approach to labelling.

Recent work \citep{wang2025doeshigherinterpretabilityimply} has suggested that high interpretability of a feature tends to have some correlation with high utility for model steering, and that this relationship seems most pronounced for JumpReLU SAEs with a low $\ell_0$ (i.e. high sparsity). This result could pose as a relevant context for interpreting the findings of \cite{wu2025axbench}, in which they base their conclusions on SAEs with high $\ell_0$, being those that are already annotated in Neuronpedia\footnote{\url{https://www.neuronpedia.org/gemma-scope\#browse}}.







\subsection{Contributions}
We make the following contributions:
\begin{enumerate}
    \item We introduce a labelling pipeline for Sparse Autoencoders based on labelled datasets rather than LLMs.
    \item We demonstrate that Sparse Autoencoders are substantially more powerful for steering than reported in \cite{wu2025axbench} when the features are labelled via our proposed method. However, the steering performance we demonstrate is still not fully competitive with prompting on the AxBench benchmark.
    \item We find indications that both low- and high-sparsity Sparse Autoencoders in Gemma Scope can be used for steering.
\end{enumerate}
\section{Preliminaries}
\label{sec:preliminaries}

\subsection{Large Language Models}
\noindent\textbf{Formal definition.} 
Large Language Models define a distribution over text strings, and are typically based on transformers \citep{vaswani2023attentionneed}. To transform a string of text into a format suitable for language modeling, the typical approach is to use a tokenizer to split up the text into a string of tokens. Here, the tokens constitute an \textit{alphabet} $\Sigma$. For modeling purposes, we further augment the alphabet by the tokens BOS and EOS (beginning of sequence, end of sequence) to get an augmented alphabet $\bar{\Sigma}=\Sigma\cup\{\textrm{BOS},\textrm{EOS}\}.$ Formally, a sequence of tokens is simply called a \textit{string}.
The language model can then be characterized as a string distribution \begin{equation}
    p_{\rm LM}(\boldsymbol{s})=\prod_i p_{\rm SM}(s_i|\boldsymbol{s}_{<i}),\quad  \boldsymbol{s}\in\left(\textrm{BOS}\cdot\Sigma^*\cdot\textrm{EOS}\right),\label{eq:lm_def}
\end{equation} where $\cdot$ denotes string concatenation and $p_{\rm SM}(\cdot|\boldsymbol{s}_{<i}):\bar{\Sigma}\rightarrow[0,1]$ is typically fitted by the model. Here $p_{\rm LM}$ is assumed to be a true distribution (i.e., no probability mass is leaked into the set of infinite strings). This is true for models with a softmax-based prediction head.
To use the language model generatively, we use a decoding strategy to auto-regressively predict the next token until termination by EOS. The two simplest options are to greedily select $\arg\max_{s_i}p_{\rm SM}(s_i|\boldsymbol{s}_{<i})$ as our next token, or to sample from $p_{\rm SM}(s_i|\boldsymbol{s}_{<i})$ at each step \citep{cotterell2024formalaspectslanguagemodeling}.

\begin{figure}
    \centering
    \includegraphics[width=\linewidth]{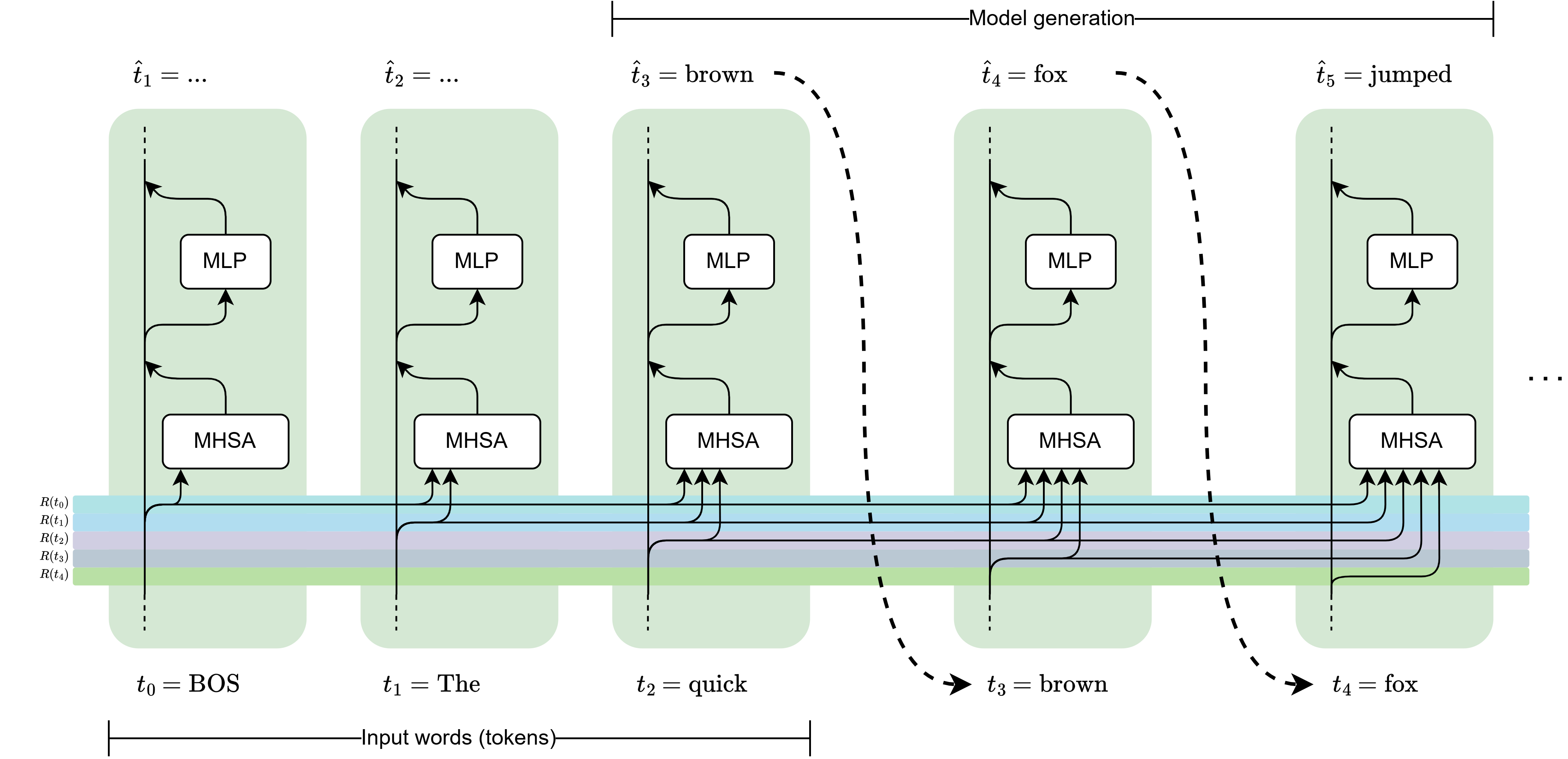}
    \caption{The architecture of a transformer-based causal LM at the level of a single transformer block. Here we see the residual stream as the vertical line running through the block, and the consequences of the causal mask on the attention. For simplicity, we assume the example words are mapped to single tokens and leave out the embedding and decoding step, as well as the positional encoding. The model generation bracket shows the autoregressive nature of the model.}
    \label{fig:transformer-llm}
\end{figure}

\noindent\textbf{Transformers.}
LLMs typically build on a modified transformer architecture compared to the one introduced in \citet{vaswani2023attentionneed}. The main modifications are to use only the decoder, to apply the positional encodings within the Multi-Head Self-Attention, and to use the causal mask.
The causal mask has the effect of making information always flow forward over a sequence of tokens.
E.g., for two sentences "The bar closes at 4" and "The bar is made of metal", the internal representation for the word "bar" cannot be disambiguated at its token position using the succeeding tokens as they are masked out. This has the advantage of being computationally cheaper since the current representations need not be recomputed every time a new token is generated. This is depicted in figure \ref{fig:transformer-llm}.
For the reader less familiar with LLMs, we will occasionally refer to the chain of residual connections that flows throughout the network as the "residual stream". This view has been widely adopted in the literature, aptly so since there is a direct path from the input layer to the output layer through it, where the transformer blocks can be seen as reading and writing its value \citep{elhage2021mathematical, lieberum2024gemma}.

%

\noindent\textbf{Prompt Steering.}
Prompting offers a straightforward way to introduce a concept into a model generation. For instance, if we want the language model to generate a poem about baseball, a natural approach is to use the prompt: "Write a poem about baseball".
In this work, we follow the procedure of \citet{wu2025axbench}. Here we ask GPT-4o-mini to write a prompt explaining our subject LLM to include a topic (e.g., "baseball") into its generation, to which we append the instruction (e.g., "Write a poem"). One such prompt could have the format:
\begin{quote}
    You are a language model that must always incorporate the concept of baseball into your responses, [...]

    Write a poem.
\end{quote}

\subsection{Sparse Autoencoders}\label{sec:sparse-autoencoders}
Sparse Autoencoders are attributed to \citet{ng2011sparse}. 
A popular interpretation of SAEs is that they learn an over-complete dictionary of features. An alternative view is as a set of linear probes without known labels, which is the view we adopt.

\noindent\textbf{Definition.}
SAEs are shallow, wide autoencoders, typically trained to reconstruct activations while promoting sparsity in the latent layer. Formally, the model can be described as
\begin{align}
    \textbf{(Encoder):}\quad&\boldsymbol{z}(\boldsymbol{x})=\sigma(\boldsymbol{x}\boldsymbol{W}_{\rm enc}+\boldsymbol{b}_{\rm enc})\label{eq:enc}\\
    \textbf{(Decoder):}\quad&\hat{\boldsymbol{x}}(\boldsymbol{z})=\boldsymbol{z}\boldsymbol{W}_{\rm dec}+\boldsymbol{b}_{\rm dec},\label{eq:dec}
\end{align}
where $\boldsymbol{x}\in\mathbb{R}^{d_{\rm model}},\hat{\boldsymbol{x}}\in\mathbb{R}^{d_{\rm model}}$, and $\boldsymbol{z}\in\mathbb{R}^{d_{\rm SAE}}$ respectively denote the network activation, the activation reconstruction and the \textit{feature activation vector}. In SAEs, we have $d_{\rm model} \ll d_{\rm SAE}$.
We can consider the encoder as discriminative/detective, and the decoder part as generative. As demonstrated in \citet{HAUFE201496}, the discriminative component may not exclusively reflect our signal of interest, hence we do not expect the columns of $W_{\rm enc}$ to be clean feature representations. Instead, due to the generative nature of the decoder, it is more natural to use the rows of $W_{\rm dec}$ as the learned feature representations in concordance with e.g. \citet{bricken2023monosemanticity,templeton2024scaling}. It is important to note that these features are not guaranteed to be an exhaustive list \citep{paulo2026sparse}.
In the case of Gemma Scope \citep{lieberum2024gemma},  the activation $\sigma(\cdot)$ is the $\textrm{JumpReLU}_{\boldsymbol{\theta}}$ function \citep[originally introduced in][]{erichson2019jumprelu}:
\begin{equation}
    \sigma(\boldsymbol{v})=\textrm{JumpReLU}_{\boldsymbol{\theta}}(\boldsymbol{v})=\boldsymbol{v}\odot \mathbf{1}\left[\boldsymbol{v}\geq\boldsymbol{\theta}\right]
\end{equation} where $\odot$ is the element-wise (Hadamard) product. 
The encoder has learnable weights $\boldsymbol{W}_{\rm enc}\in\mathbb{R}^{d_{\rm model}\times d_{\rm SAE}},$ learnable bias $\boldsymbol{b}_{\rm enc}\in\mathbb{R}^{d_{\rm SAE}}$, and a learnable threshold $\boldsymbol{\theta}\in\mathbb{R}^{d_{\rm SAE}}$.
The decoder has the learnable weights $\boldsymbol{W}_{\rm dec}\in\mathbb{R}^{d_{\rm SAE}\times d_{\rm model}},$ and the learnable bias $\boldsymbol{b}_{\rm dec}\in\mathbb{R}^{d_{\rm model}}$. 
Together, the model is trained to minimize $\mathcal{L}=||\boldsymbol{x}-\hat{\boldsymbol{x}}||_2^2+\lambda ||\boldsymbol{z}(\boldsymbol{x})||_0$, where the second term incentivizes sparsity on feature activations \citep{lieberum2024gemma,bricken2023monosemanticity,gao2024scaling,rajamanoharan2024jumping}.
For a technical explanation on how this is performed in practice, refer specifically to \citet{rajamanoharan2024jumping}.

\noindent\textbf{Feature Steering.}
In this work, we use the approach of \citet{templeton2024scaling} for activation editing.
During inference, we clamp the activation of a feature $\left(\boldsymbol{z}(\boldsymbol{x})\right)_f$ to a value $\alpha$. The specific approach first computes the reconstruction error term
\begin{equation}
\boldsymbol{e}=\boldsymbol{x}-\hat{\boldsymbol{x}}(\boldsymbol{z}(\boldsymbol{x})),
\end{equation}
followed by the latent vector after intervention via editing;
\begin{equation}\label{eq:intervention}
    \boldsymbol{z}_{\rm int}=\boldsymbol{z}(\boldsymbol{x})\odot (1-\boldsymbol{m})+\alpha\boldsymbol{m}
\end{equation}
where $\boldsymbol{m}$ is a mask consisting of elements $m_j=\mathbf{1}\left[j=f\right].$
The edited activations can finally be computed as \begin{equation}
    \hat{\boldsymbol{x}}_{\rm int}=\hat{x}(\boldsymbol{z}_{\rm int})+\boldsymbol{e},
\end{equation} which we pass through the rest of the model. 
This is done on all tokens, including the BOS token. We shall refer to this procedure as \textit{feature steering}. 

\section{Method}
\subsection{Labelling pipeline}\label{sec:pipeline}
To label the learned features of an SAE, we propose to construct linear probes based on single features and match their predictions with labels on a multi-label dataset. Here we use the LLM activations $\tens{X}\in\mathbb{R}^{|\mathcal{D}|\times T\times d_{\rm model}}$ computed over a dataset $\mathcal{D}$ with maximum sequence length $T$, as well as a binary mask $\boldsymbol{M}\in\{0,1\}^{|\mathcal{D}|\times T}$ zeroing out token positions not included in a given sample. The dataset comes with a binary label matrix $\boldsymbol{Y}\in\{0,1\}^{|\mathcal{D}|\times n_{\rm labels}}.$

\noindent\textbf{Feature probes.} We construct the probe predictions by first computing the feature activations $\tens{Z}_{i,j,:}=\boldsymbol{z}\left(\tens{X}_{i,j,:}\right)$ resulting in a sparse tensor $\tens{Z}\in\mathbb{R}^{|\mathcal{D}|\times T\times d_{\rm SAE}}.$
Given the activations of feature $f$, we then compute its activation frequency on the $i$'th text example:
\begin{equation}
    \boldsymbol{F}_{i,f}=\left(\sum_{j=1}^T \boldsymbol{M}_{i,j}\right)^{-1}\sum_{j=1}^T \boldsymbol{M}_{i,j}\cdot1\left[\boldsymbol{Z}_{i,j,f}>0\right]\quad\in[0,1],
\end{equation} using the indicator function $1[\cdot]$.
We define $\hat{y}_{i,f,\tau}=1\left[F_{i,f}>\tau\right]$ to form the probe prediction given a threshold $\tau$. Since it is not obvious what would constitute a good frequency threshold, we let $\tau$ be a free parameter.


\noindent\textbf{Probe labelling.} To match a dataset label to a probe, we compute a match score using the calibrated F1 \citep{siblini2020} followed by additional denoising. For every feature-label pair $(\ell, f)$, we define the calibrated F1 score as:
\begin{align}\label{eq:calibrated_f1}
    \textrm{TP}_{\ell,f,\tau}&=\sum_{j=1}^{|\mathcal{D}|}\mathbf{1}\left[y_{j,\ell}=1\,\wedge\,\hat{y}_{j,f,\tau}=1\right],\quad
    &\textrm{FP}_{\ell,f,\tau}&=\sum_{j=1}^{|\mathcal{D}|}\mathbf{1}\left[y_{j,\ell}=0\,\wedge\,\hat{y}_{j,f,\tau}=1\right],\\
    \textrm{FN}_{\ell,f,\tau}&=\sum_{j=1}^{|\mathcal{D}|}\mathbf{1}\left[y_{j,\ell}=1\,\wedge\,\hat{y}_{j,f,\tau}=0\right],\quad
    &\pi_{\ell}&=\frac{1}{|\mathcal{D}|}\sum_{j=1}^{|\mathcal{D}|}1\left[y_{j,\ell}=1\right],\\
    \textrm{F1}^c_{\ell,f,\pi_0}&:=\max_{\tau}\left[\frac{2\textrm{TP}_{\ell,f,\tau}}{2\textrm{TP}_{\ell,f,\tau}+\frac{(1-\pi_0)\pi_{\ell}}{\pi_0(1-\pi_{\ell})}\textrm{FP}_{\ell,f,\tau}+\textrm{FN}_{\ell,f,\tau}}\right].
\end{align}
We further explore this definition in appendix \ref{appendix:calibrated_f1} and examine the effects of the parameter $\pi_0.$
In practice, the max operation selects the optimal threshold $\tau$ over a set of candidates $\{0.05,0.1,...,0.95\}.$

To motivate the use of the F1 score, consider its decomposition into the harmonic mean between precision and recall, i.e. $\textrm{F1}=2\left(\textrm{precision}^{-1}+\textrm{recall}^{-1}\right)^{-1}.$ The objective is for a probe to have both high precision (almost exclusive activation) and high recall (activation on most occurrences) for the correct label. A probe scoring highly in one but low in the other would likely correspond to a super- or a sub-category, or something semantically adjacent. We here took inspiration from insights of \citet{gao2024scaling, gurnee2023finding}. 
To motivate the calibration, we consider how the ordinary F1 score is sensitive to the label \textit{support} (the number of occurrences for the label). Since labels may have different supports, the scale of the F1 score will vary between labels, rendering it unreliable for comparing the degree of agreement between a feature and a label. Again, we refer the reader to appendix \ref{appendix:calibrated_f1} for a detailed exploration.

To select a set of valid matches based on $\textrm{F1}^c_{\ell,f,\pi_0}$, we devise a set of criteria to \textit{denoise} the result. These are performed in the following order:\begin{enumerate}
    \item To reduce variance, we threshold the label support, i.e., the number of label occurrences in the dataset. We found $50$ to be a good threshold through parameter tuning, which we elaborate in section \ref{sec:experiment}.
    \item We use the output score from \citet{arad2025saesgoodsteering} as a proxy for the steering ability of a feature. We use a threshold of $10^{-3}$ found by visually inspecting the output score distribution and choosing it to only include a small best subset.
    \item We keep the top-K of the remaining feature-label pairs. We use $K=50$, since it provides a tradeoff between accuracy and number of matches.
\end{enumerate}

\subsection{AxBench}\label{sec:axbench}
AxBench is a benchmark for model generation under steering, prompting the model with Alpaca-Eval instructions \citep{wu2025axbench,alpaca_eval}.
The generated texts are rated by GPT-4o with scores $\{0,1,2\}$ on 1) fluency (whether they are correct English); 2) instruction following (whether they respond to the given instruction); and 3) concept incorporation (whether they naturally include the target concept). The scores are finally combined into a single overall score for each generation by the harmonic mean $$\textrm{aggregated\_rating}=\frac{3}{\textrm{instruction\_rating}^{-1} + \textrm{concept\_rating}^{-1}+\textrm{fluency\_rating}^{-1}}.$$ The ordinary mean is used to aggregate each score over the dataset.

\noindent\textbf{Feature Steering Evaluation.}
To generate a set of steered texts, we first draw 20 random instructions from Alpaca-Eval without replacement. We repeat this for each suggested feature-label pair. For a set of considered steering strengths $\alpha\in\{100,125,...,575\}$, we generate feature-steered responses to the selected instructions using each strength value.

In \citet{wu2025axbench}, they select $\alpha$ by maximizing $\textrm{aggregated\_rating}$ in holdout (5 training examples, 5 test examples). The resulting value is used across all features.
However, since we do not expect all features to require the same $\alpha$ for optimal steering, we select it individually for each feature-label pair. Here, we generate 20 responses for every considered $\alpha$, for every suggested feature-label pair. We then report the result from repeated cross-validation (10 repeats, 4 splits) to reduce variance\footnote{For each split, we select $\alpha$ by maximizing over the training-partition and report the score on the evaluation-partition}.
To evaluate the overall steering, we use $\textrm{aggregated\_rating}$ as the maximization objective, as in \citet{wu2025axbench}. However, since the aggregated rating accounts for model-specific abilities, we use the $\textrm{concept\_rating}$ to evaluate the accuracy of our feature-labels.

To offer an additional motivation for a steering-based pipeline evaluation, this approach avoids the potential issue of dataset idiosyncrasy that may cause an interpretability illusion, as observed in \citet{bolukbasi2021interpretability}. Here, the observed semantics of a neuron vary across the datasets used to identify them, possibly explained by neuron polysemanticity. In contrast, feature steering is a causal experiment, and while it additionally demands causation, a successful feature steering would confirm what we may consider the "ground truth" labels of the learned features.

\noindent\textbf{Prompt Steering Evaluation.}
For each feature-label pair, we use the same 20 random instructions. We then generate the prompt-steered responses to each instruction. Since there are no parameters to select for prompting, we report the average.

\subsection{Resources}
\noindent\textbf{Datasets.} As our multi-label dataset, we use Stack Exchange obtained by the tool of Eleuther-AI to process the data dump on the Internet Archive \citep{stackexchange-eleutherai, stackexchange-datadump}.
As labels, we use the tags assigned to a post by the user -- for instance, a user may assign "baseball" and "mlb" (Major League Baseball) to a post about baseball. 
As our text inputs, we clean the post and append the comments made by other users. Since the data format is a set of large XML files, we create an index of all newline characters in the bytestream to achieve fast random access. 
Since there are multiple Stack Exchange fora available, we obtain a suite of datasets related to different topics. We chose the following: academia, biology, chemistry, cooking, cs, history, law, literature, physics, politics, and sports. For each forum, we plot the cumulative density functions of label supports after pre-processing in figure \ref{fig:label-support-cdf}.

\noindent\textbf{Models.} We run the experiments on Gemma 2 \citep{team2024gemma} + Gemma Scope \citep{lieberum2024gemma}. As a baseline for feature labels, we use the ones provided in Neuronpedia.

\noindent\textbf{Hardware.}
The entire experiment can be run on a single H100 node within 24 hours for most Stack Exchange fora. 
To properly control the GPU memory consumption, we have used a set of code optimizations, including pre-allocation and in-place computation of most matrices, and an OOM recovery mechanism that successively halves the batch size if needed. 
The codebase was originally developed on an RTX4090 machine, and the full pipeline can be run on this GPU for some Stack Exchange fora.
\section{Experimental design and results}\label{sec:experiment}
\noindent\textbf{Setup.} Using the selected Stack Exchange fora, we apply the pipeline to two different layers in the model for completeness. Here, we use layer 17 and layer 32.\footnote{We enumerate the layers using the original indices from \cite{lieberum2024gemma} starting from zero.}
Our method was originally developed and tuned using layer 32, which we chose in the early experimental phase for Gemma-2-9b to reside in the second half of the model, anticipating reasonable steering performance. 
We arbitrarily selected layer 17 to examine whether our method would also generalize to earlier layers. Layer 17 was excluded from method development and parameter tuning to ensure a reliable result.

\noindent\textbf{Parameter tuning.} 
Since our pipeline contains a set of parameters ($\pi_0$, $K$, support threshold, and output score threshold), we use a hold-out approach to tuning. Here, we exclusively tune using layer 32 and the fora: academia, biology, history, law, literature, physics, and sports. Hence, the remaining fora and layer 17 are not included. 
As elaborated in section \ref{sec:pipeline}, $K$ and the output score threshold were chosen heuristically in the early experimentation in this setup.
We find that pipeline performance seems particularly sensitive to the choices of $\pi_0$ and the support threshold. Since it is infeasible to evaluate it on the full Cartesian product of $\textrm{features}\times\textrm{labels},$ we use a manual iterative approach. Here, we select sensible initial values and evaluate the assigned labels. We then inspect the results, tune the parameters optimizing for $\textrm{concept\_rating}$, and repeat. After a few iterations, we converge at $\pi_0=10^{-3}$ and a support threshold of 50, still being sensible values.

\noindent\textbf{Baseline.} As baselines, we use those provided in \cite{wu2025axbench}, including a reproduction of the SAE and prompt baselines named NP random. NP random samples 100 features at random and evaluates their official Neuronpedia labels for steering and prompting.

\noindent\textbf{Interpretation.} Comparisons of steering performances between feature-steering and prompt-steering are shown in figures \ref{fig:layers-aggregated} and \ref{fig:layers-concept} (see appendix \ref{appendix:plots}). 
In figure \ref{fig:layers-concept}, we see that our proposed pipeline is indeed capable of assigning correct labels to the causal effects of features. While we observe some variability between different fora, we mostly observe similar effects when comparing feature-steering to prompt-steering.

Figure \ref{fig:layers-aggregated} depicts the overall steering performance and shows a general improvement over those reported in \citet{wu2025axbench}. We observe that multiple fora provide labels that result in feature-steering performance on par with the LoRA reference.
We also observe a clear improvement in the prompt-steering performance, possibly due to the subject model's better understanding of simpler labels. The reference Neuronpedia performance is largely consistent with our results.

When considering the performance increase of both feature-steering and prompt-steering from the Neuronpedia reference, one could be tempted to alternatively suggest better label comprehension by the LLM judge as an explanation. However, when considering the $\textrm{concept\_rating}$ performance in figure \ref{fig:layers-concept}, we now observe a mostly closed gap between feature-steering and prompt-steering compared to the Neuronpedia baseline. Since $\textrm{concept\_rating}$ is the only component in AxBench that relies on the feature-label, the alternative explanation does not fully account for the results.

\begin{sidewaysfigure}[p]
    \centering
    \includegraphics[width=\linewidth]{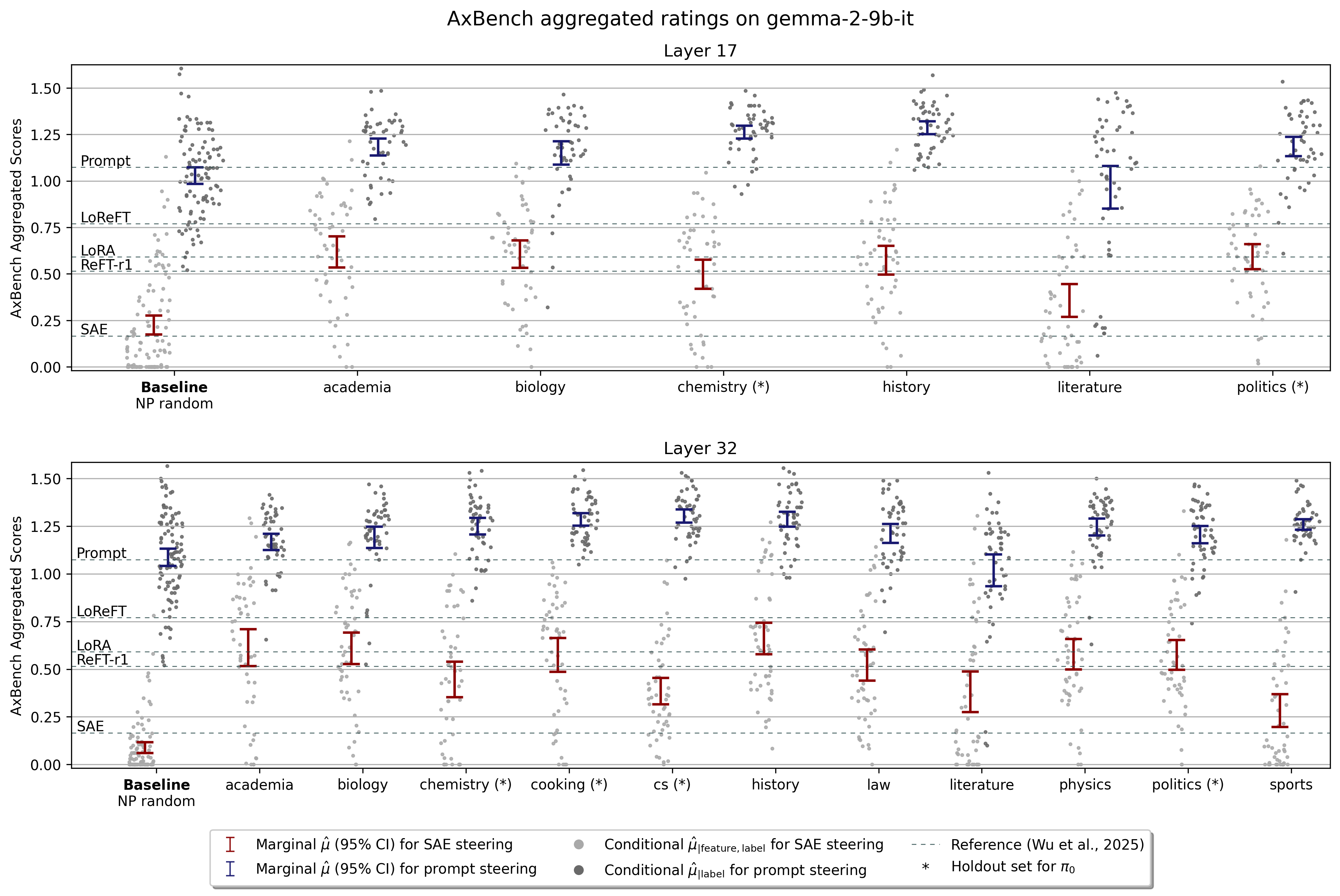}
    \caption{AxBench aggregated ratings for two layers of Gemma-2-9b-it across multiple stackexchanges. Here the Gemma-Scope SAEs have width 131k, are inserted into the residual stream, and have sparsity parameters $\ell_0\approx11$ and $\ell_0\approx10$ respectively (highest sparsity). The gray datapoints illustrate the average performances of individual selected label, where the jitter on the x-axis is illustrative of the density estimated via KDE. The confidence intervals were computed using non-parametric bootstrapping.
    The figure also shows multiple baselines: NP random being 100 randomly selected feature-label pairs from Neuronpedia, and multiple performances from \cite{wu2025axbench}. The Neuronpedia baselines are based on the higher $\ell_0$ counterparts of the selected SAEs.}
    \label{fig:layers-aggregated}
\end{sidewaysfigure}

\noindent\textbf{Ablation.} We perform an ablation study to examine the impacts of the roles of 1) a high sparsity, and 2) the output score. The results are shown in figures \ref{fig:ablation-aggregated} and \ref{fig:ablation-concept}.
Here we see that using high sparsity SAEs (i.e., low $\ell_0$) seemingly does not greatly impact the steering performance.
We also observe a rather modest performance improvement caused by the output score, in both the low and high sparsity SAEs. 

To examine the role of the output score, we compare randomly selected features with features selected by our pipeline. In both feature sets, we fetch their original Neuronpedia labels.
Here, we observe better feature-steering performance of those selected by the pipeline, even without using the output score. 

\section{Discussion}
We find the results of section \ref{sec:experiment} somewhat surprising for the two reasons: 1) causality generally cannot be taken for granted given a correlation/dependence, yet removing the causal-component of the pipeline (the output-score) does not greatly affect its performance; and 2) the reported feature-steering performance in \cite{wu2025axbench} was much weaker compared to our setup.

\noindent\textbf{Linear Representation Hypothesis.} While \cite{wu2025axbench} reported obtaining rather poor steering performance on all methods that linearly steer, we found that SAEs can, surprisingly, steer with a performance close to par with LoRA that uses adapter weights \citep{hu2022lora}. This finding demonstrates the power of using linear methods for understanding the internals of language models.


\noindent\textbf{AxBench.} Since AxBench uses an LLM-judge to rate steered generations, we find it natural to ask whether the LLM-judge could exhibit a bias to favor some steering methods differently than a population of human judges.
To investigate this, we conducted a small trial with Danish bachelor students and found indications that they would generally rate some scores more optimistically than GPT-4o-mini, however, we did not find clear evidence of a bias in the trial. We note that the trial had a small sample size, so a more thorough user study might still be relevant. 
However, while inspecting our pipeline evaluation results, we observed multiple examples of a label \textit{semantically adjacent} to, or even fully matching, the elicited topic, where AxBench still assigned it a concept score of 0. We included some of these examples in appendix \ref{sec:appendix-3}.
Since AxBench relies on simply prompting the LLM judge with a detailed instruction without examples, it seems natural to ask whether a few-shot prompting approach could benefit its accuracy and alignment.

\noindent\textbf{Applications.} Although our results suggest an improved feature-steering performance compared to \cite{wu2025axbench}, it is important to acknowledge that prompting is a baseline that is likely hard to match, as the model was trained to respond to prompts. In our view, feature-steering is an exercise in \textit{controlled} damage of the subject model, but it is a damage nonetheless, rendering it less useful in production.
However, in a research setting, the causal nature of features can be reassuring that an assigned label is indeed the ground truth, and not simply an artifact of an idiosyncratic dataset \citep{bolukbasi2021interpretability}. Hence, Sparse Autoencoders could still be of interest for interpretability and explainability researchers. Our proposed supervised pipeline also has the advantage of being applicable outside the text-domain, potentially enabling researchers to explore the internals of models in computer vision, EEG-data, etc.
\noindent\textbf{Influence of support.} Since we found it beneficial to use a relatively high support threshold, it would be reasonable to reflect on its effects on the setup. While its primary motivation was to reduce variance in the calibrated F1 score (a direct effect), it likely also indicates the general popularity of a topic outside the respective Stack Exchange forum. Since it is not public knowledge whether Gemma-Scope was trained on Stack Exchange, we can only speculate that topics of higher general popularity may also more frequently occur in its training set. 
As the reconstruction loss will likely favor more frequent topics to be learned as features \citep{muhamed2025decoding}, the high support threshold could filter out unlikely candidate labels as an indirect effect.






\section{Conclusion}
Sparse Autoencoders initially seemed promising for exploring the internals of LLMs, and gained traction following publications such as \cite{bricken2023monosemanticity,templeton2024scaling,lieberum2024gemma} and more. However, the optimism around the model has since faded, and its reliability has been cast into question. In this work, we have demonstrated that Sparse Autoencoders can, to an extent, live up to the original hype, even if they are not the panacea of interpretability and steering once thought to be. 
We find that interpretable features of SAEs exhibit a surprisingly causal impact on the model predictions when the labels are selected by a different approach than the one used to construct the Neuronpedia labels. Our proposed labelling method allows Sparse Autoencoders to perform on par with the LoRA baseline provided in \cite{wu2025axbench}, vastly outperforming their reported capabilities. 
We further see indications that both low- and high-sparsity Sparse Autoencoders can be leveraged for steering tasks.
\subsubsection*{Broader Impact Statement}
This work explores Sparse Autoencoders for model steering and interpretability. We believe the direct ethical and societal implications of our work are minimal, as it primarily contributes to an understanding of language models. In a broader perspective, we believe this could be useful for performing safety auditing, model debugging, and alignment research. We acknowledge the potential risk associated with enabling direct manipulation of the open source models that could enable the expression of possessed dangerous capabilities. However, we consider the additional risk posed by our framework insignificant compared to standard finetuning approaches.

\makeatletter
\if@accepted
\subsubsection*{Acknowledgements}
Special thanks to Hiba Nassar at DTU Cognitive Systems for letting us run the user study in her course "02462 Signals and Data" at DTU Compute in Autumn 2025. This work was supported by the Novo Nordisk Foundation grant NNF22OC0076907 “Cognitive spaces - Next generation explainability” and by the Pioneer Centre for AI, DNRF grant number P1.
\fi
\makeatother

\bibliography{bibliography.bib}
\bibliographystyle{tmlr}

\appendix
\clearpage
\section{Appendix: Calibrated F1}\label{appendix:calibrated_f1}
\subsection{Unpacking the calibrated F1 score}
To simplify the notation, we define the events $+$ and $-$ as the positive and negative classes of the label, and let $\hat{+}$ and $\hat{-}$ define the events of positive and negative predictions. We also abuse the notation and consider $\pi_0$ a Bernoulli distribution over $\{+,-\}$. 

A quick glance at the F1 score reveals its dependency on the class prior: \begin{equation}
    \begin{split}
        \textrm{F1}&=\frac{2\textrm{TPR}}{2\textrm{TPR}+\textrm{FPR}+\textrm{FNR}}\\
        &=\frac{2P(\hat{+},+)}{2P(\hat{+},+)+P(\hat{+},-)+P(\hat{-},+)}\\
        &=\frac{2P(\hat{+}|+)}{2P(\hat{+}|+)+\frac{P(-)}{P(+)}P(\hat{+}|-)+P(\hat{-}|+)}.
    \end{split}
\end{equation}

We can fix this by using the calibrated F1 of equation \ref{eq:calibrated_f1}:\begin{equation}
    \begin{split}
        \textrm{F1}^c_{\pi_0}&=\frac{2\textrm{TPR}}{2\textrm{TPR}+\frac{\pi_0(-)P(+)}{\pi_0(+)P(-)}\textrm{FPR}+\textrm{FNR}}\\
        &=\frac{2P(\hat{+}|+)}{2P(\hat{+}|+)+\frac{\pi_0(-)}{\pi_0(+)}P(\hat{+}|-)+P(\hat{-}|+)}=\textrm{F1}_{y\sim \pi_0}.
    \end{split}
\end{equation}
Here we see that any change to the true fraction $\frac{P(-)}{P(+)}$ will not impact the calibrated F1 score. Hence, it is invariant to the true class prior/support, and computes the F1 under the provided prior $\pi_0$.

\subsection{The impact of $\pi_0$ in our setup}
At first glance, it could seem tempting to set $\pi_0=\frac{1}{2}$ to make the influences of $\textrm{FP}$ and $\textrm{FN}$ equal. However, we found that this does not work well in practice, and that, in fact, the labeling pipeline benefits tremendously from having a low $\pi_0$. 
We have a hypothesis for why that is the case: On stackexchange, when a user writes a post, they typically select a set of tags describing it. Here they may be more prone to excluding a topic in the tag-line than to include tags that are irrelevant to the post. For instance: A user might briefly mention \textit{Europe} in a post tagged with \textit{PhD} and \textit{Applications}, whence \textit{Europe} becomes a secondary topic in the post but not a primary one. Conversely, a user might be less prone to writing a post exclusively about e.g. \textit{baseball} and then to tag it with e.g. \textit{Formula-1}. 
Here the first case would result in a \textit{false positive} when considering the tag - the post does, in fact, include \textit{Europe} but the tagline does not. Similarly, the second case gives a \textit{false negative} because the post does not include Formula-1 although the tagline suggests so.
Since these tags constitute our labels, we expect there to be more false positives (w.r.t. the assigned labels) than false negatives in the dataset. We further expect the secondary topics to occur with a lower frequency within a single post than the main topics. 
This could motivate selecting a higher threshold in the probe to suppress the competition among labels.

In the following, we demonstrate in a simulation how setting $\pi_0$ to a low value will provide these benefits.

\subsubsection{Simulation}\label{sec:simulations}
\begin{figure*}[t!]
    \centering
    \begin{subfigure}[b]{0.5\textwidth}
        \centering
        \includegraphics[width=\linewidth]{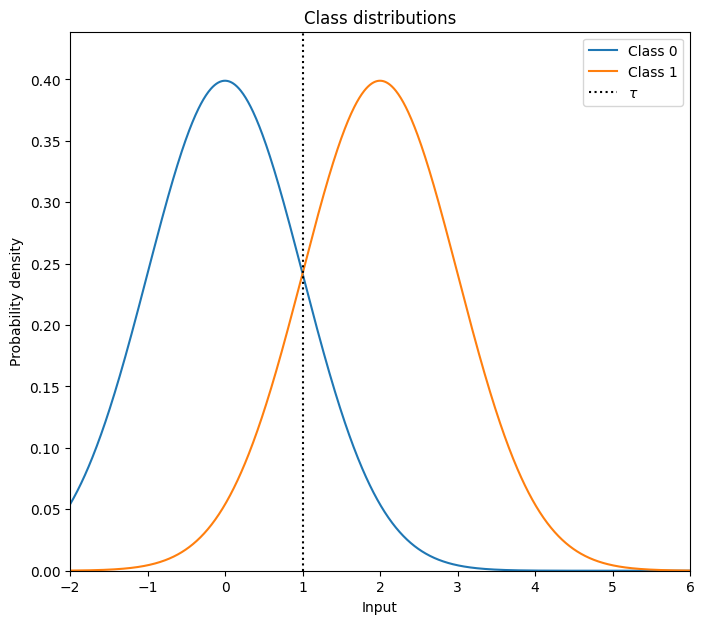}
        \caption{}
        \label{fig:f1-simulation-distributions}
    \end{subfigure}%
    ~ 
    \begin{subfigure}[b]{0.5\textwidth}
        \centering
        \includegraphics[width=\linewidth]{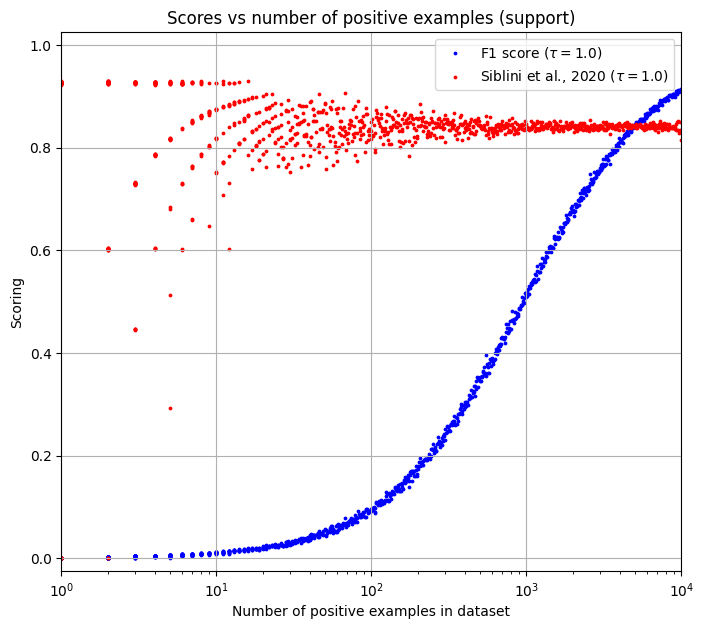}
        \caption{}
        \label{fig:f1-simulation-instability}
    \end{subfigure}
    \caption{
        a) The distributions $\pi_0=\mathcal{N}(\mu_1=0,\sigma^2=1^2)$, and $p_1=\mathcal{N}(\mu_2=2,\sigma^2=1^2)$ used in the simulation of the calibrated F1 score.
        b) Despite the invariance of the calibrated F1 to label support, the statistic may exhibit very high variance, in a somewhat peculiar pattern, given a low number of positive samples.
    }
\end{figure*}

\begin{sidewaysfigure}[p]
    \centering
    \includegraphics[width=\textheight]{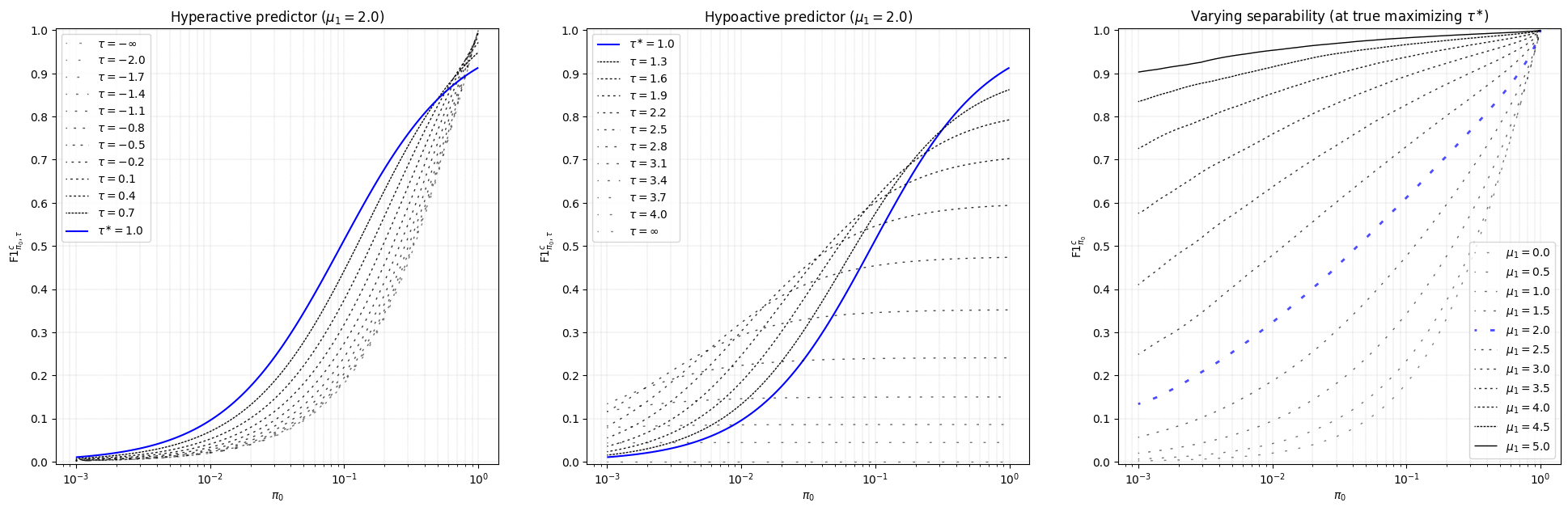}
    \caption{The simulation results of section \ref{sec:simulations}, where we compute the empirical calibrated F1 scores as a function of $\pi_0$ while sampling from $\pi_0=\mathcal{N}(\mu_1=0,\sigma^2=1^2),$ and $p_1=\mathcal{N}(\mu_2=2,\sigma^2=1^2)$. The first two figures show simulated hyperactive/hypoactive probes, i.e., probes with too many/too few positive predictions. The last figure shows the scenario where the threshold $\tau$ is set to the value maximizing $\textrm{F1}^c_{\pi_0,\tau}$, where the value of $\mu_1$ and $\pi_0$ are varied. 
    }
    \label{fig:f1_simulation}
\end{sidewaysfigure}

To simulate the setup of having a feature probe with a known ground-truth label, we sample $n$ points from two normal distributions $\pi_0=\mathcal{N}(\mu_1=0,\sigma^2=1^2),p_1=\mathcal{N}(\mu_2=2,\sigma^2=1^2)$ with a class-prior. This constitutes our $X\in\mathbb{R}^n$ (feature activity) and $Y\in\{0,1\}^n$ (class-labels). Figure \ref{fig:f1-simulation-distributions} illustrates the distributions.

We first take a look at the stability of the calibrated F1 score when there is low support. Here we vary the prior, $P(+)$, while computing the score from $n=10\,000$ data points. We see the calibrated F1 will tend to exhibit high variance in a somewhat unusual pattern for a low (absolute) number of positive samples. The apparent lines seen in the approximate interval $\#(+)\in[1, 40],$ may be explained by a combinatorial limitation, since TP and FN in this range have a small number of possible values to assume. Meanwhile, $\textrm{FP}^c_{\pi_0}=\frac{\pi_0(-)P(+)}{\pi_0(+)P(-)}\textrm{FP}$ may tend to take very low values unless $\pi_0$ corrects for it. In this figure, we set $\pi_0(+)=\frac{1}{2}.$

We then compute the calibrated F1 for a set of thresholds as a function of $\pi_0$:
$$
{\textrm{F1}}^c_{\pi_0,\tau}=\frac{2\textrm{TP}_{\tau}}{2\textrm{TP}_{\tau}+\frac{\pi_0(-)P(+)}{\pi_0(+)P(-)}\textrm{FP}_{\tau}+\textrm{FN}_{\tau}},
$$
which is shown in figure \ref{fig:f1_simulation}. Here we sample $n=1\,000\,000$ data points with a flat prior $P(+)=P(-)=\frac{1}{2}.$
To declutter the figures, we separate the thresholds below and above $1$ into the categories "hyperactive" and "hypoactive" probes. Here, a hyperactive probe will have a lower threshold than what would seem optimal given the distributions (consider shifting the threshold line to the left in figure \ref{fig:f1-simulation-distributions}), and a hypoactive probe will have a higher threshold.
In the last subfigure of figure \ref{fig:f1_simulation}, we also vary $\mu_2$ and use the maximizing threshold for each value to see how $\pi_0$ influences the requirement of separability between distributions. This is to simulate the procedure in full, where the maximizing threshold is used to compute $F^c_{\pi_0}$ as in equation \ref{eq:calibrated_f1}.

As a result of the simulations, we note the asymmetry between hypoactive and hyperactive probes at the lower values of $\pi_0.$ Here, hyperactive probes are penalized while hypoactive probes are rewarded. Furthermore, the preferred threshold causes only the examples from the higher end of the positive distribution to be classified as positive, while passing many fewer of the negative class examples. Hence, it focuses only on the part of the more reliable part of the signal (e.g., more mentions of a topic in a post), which causes fewer FPs but more FNs. 
We also see that lower values of $\pi_0$ will have a stronger preference towards higher separability.

\clearpage
\section{Appendix: Extra figures}\label{appendix:plots}

\begin{figure}[ht]
\centering
\rotatebox{90}{
    \begin{minipage}{0.9\textheight}
        \centering
        \includegraphics[width=\linewidth]{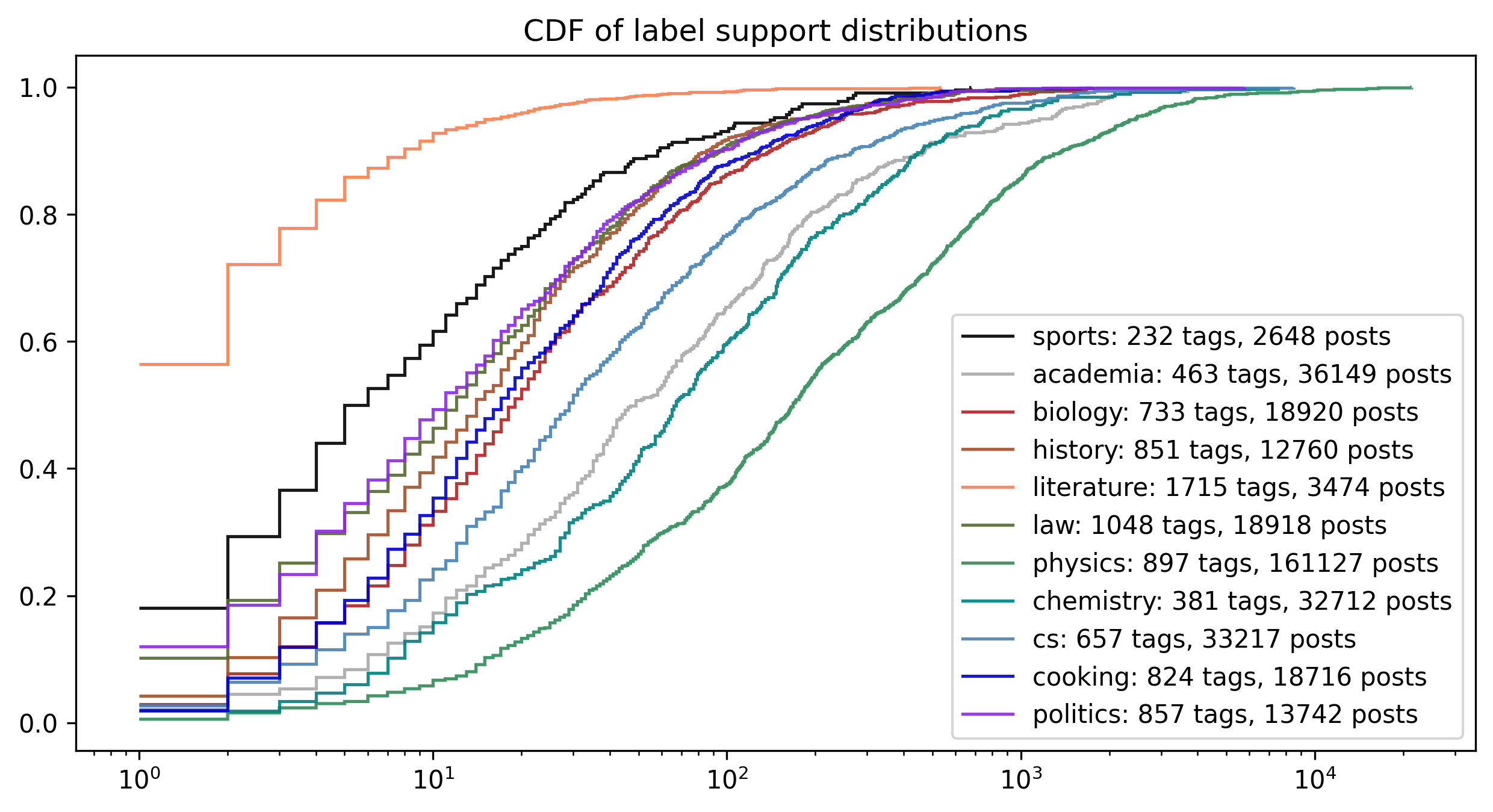}
        \captionof{figure}{Empirical Cumulative Density Functions of the label supports for the selected Stack Exchange fora after pre-processing.}
        \label{fig:label-support-cdf}
    \end{minipage}
}
\end{figure}

\begin{sidewaysfigure}[p]
    \centering
    \includegraphics[width=\linewidth]{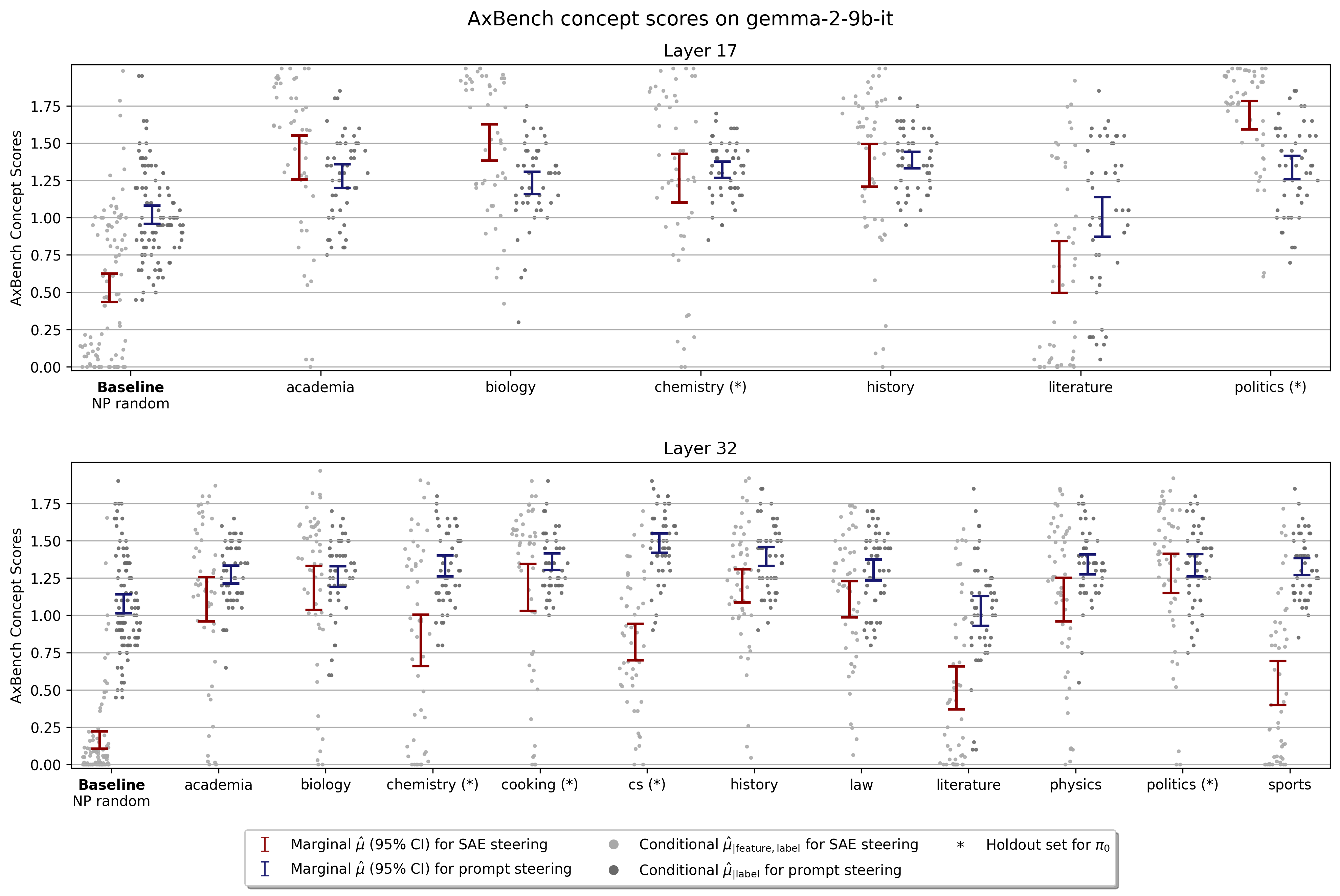}
    \caption{AxBench concept ratings for two layers of Gemma-2-9b-it across multiple fora. Here, the setup is identical to that in figure \ref{fig:layers-aggregated}.}
    \label{fig:layers-concept}
\end{sidewaysfigure}

\begin{sidewaysfigure}[p]
    \centering
    \includegraphics[width=\linewidth]{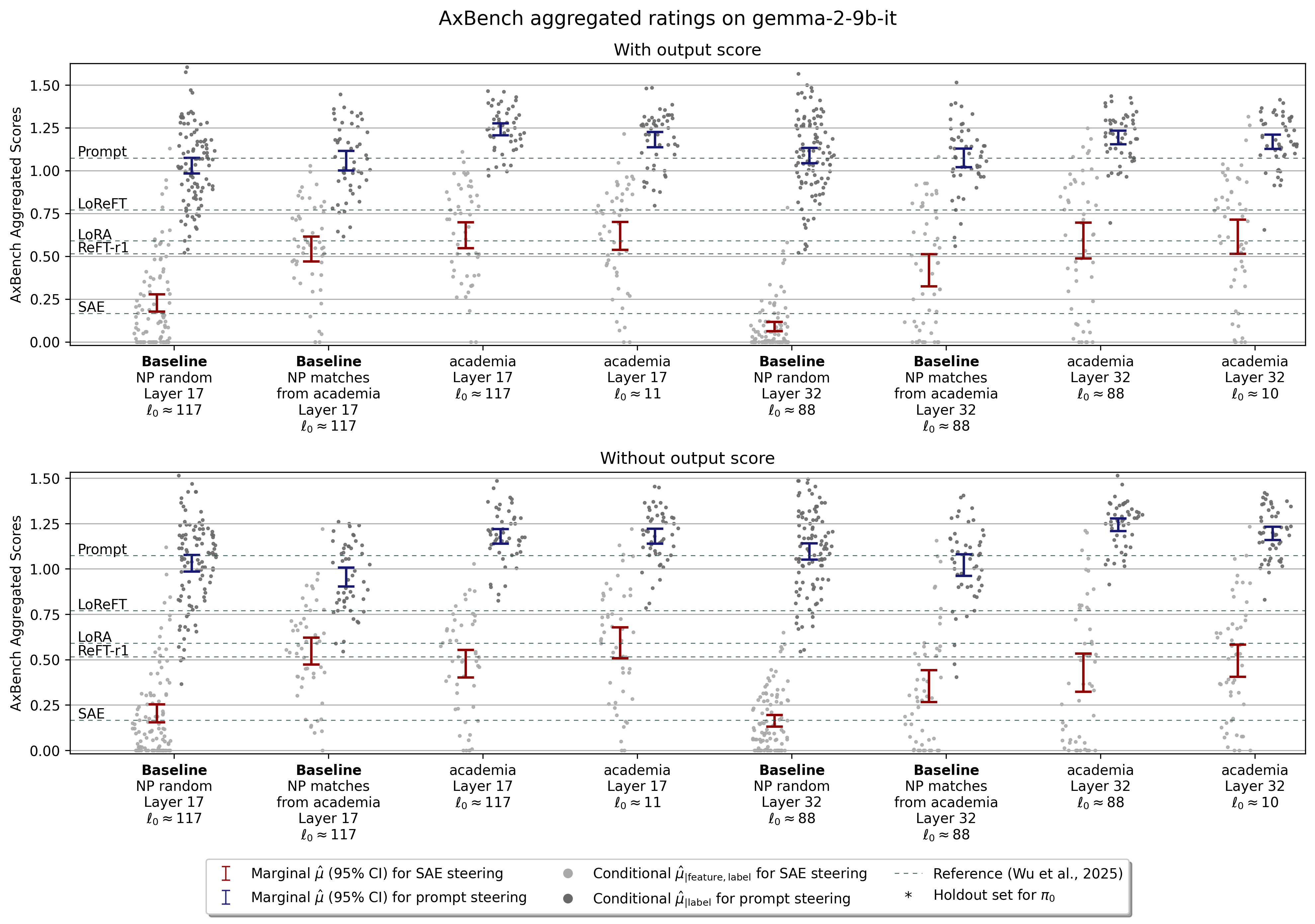}
    \caption{AxBench aggregated ratings for two layers of Gemma-2-9b-it in an ablation study. Here, the setup is identical to that in figure \ref{fig:layers-aggregated} except where stated otherwise.}
    \label{fig:ablation-aggregated}
\end{sidewaysfigure}

\begin{sidewaysfigure}[p]
    \centering
    \includegraphics[width=\linewidth]{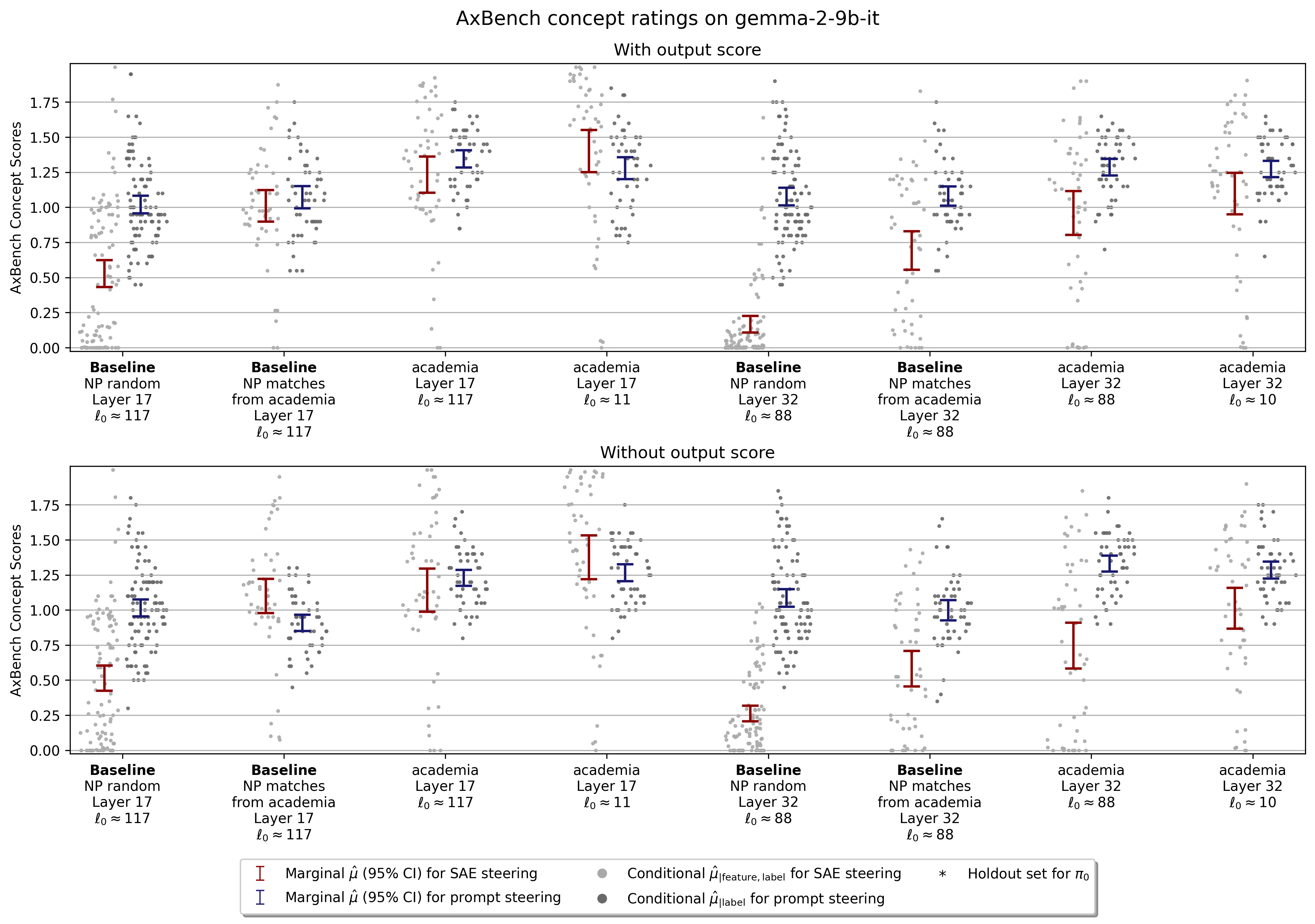}
    \caption{AxBench concept ratings for two layers of Gemma-2-9b-it in an ablation study. Here, the setup is identical to that in figure \ref{fig:layers-aggregated} except where stated otherwise.}
    \label{fig:ablation-concept}
\end{sidewaysfigure}

\clearpage
\section{Qualitative examples of steering}
To ensure that the AxBench benchmark does not exhibit any obvious flaws, we here present some qualitative examples of the model steering. The examples were picked at random, but the strengths were chosen to ensure a representative visible effect of the intervention (regardless of whether the effect matches the label). For transparency, all model generations are included in the accompanying GitHub repository in CSV-format.

\begin{tcolorbox}[title=The "China" feature - Layer 32]
\textbf{Prompt:} What are places to visit in Montgomery, Alabama?

\textbf{Response:}
\begin{lstlisting}
##  A Visit to China's Capital: Exploring Beijing

China's capital, Beijing, offers a captivating blend of ancient history and modern dynamism. 

Here are some must-visit places in Beijing:

**Historical Treasures:**

* **The Forbidden City:** This vast imperial palace complex, once home to Chinese emperors, boasts intricate architecture, grand halls, and ancient treasures.
* **The Temple of China's Heaven:** An impressive Taoist temple complex offering panoramic city views and insights into Chinese culture and philosophy.
* **The Great Wall of China (Beijing section):** Witness the world's largest man-

\end{lstlisting}

\textbf{Ratings:} fluency = 1, concept = 2, instruction = 0, aggregated = 0.
\end{tcolorbox}

\begin{tcolorbox}[title=The "Poster" feature - Layer 32]
\textbf{Prompt:} Write a symphony concert review, discussing the orchestra's performance and overall audience experience.

\textbf{Response:}
\begin{lstlisting}
## A Night of Posters and Powerhouses: Symphony Orchestra Unleashes Raw Emotion

Poster art depicting a soaring, spectral figure against a deep crimson backdrop promised a night of intense, evocative posters – and the Symphony Orchestra delivered in spades. Under the masterful guidance of conductor [name of conductor], the performance journeyed through a landscape of bold orchestral posters, showcasing the legendary lineup of musicians within this renowned organization.

The concert opened with [name of poster, composer], a visually striking portrayal of [brief description of poster's mood or theme]. The orchestra's vibrant posters were immediately apparent, showcasing the poster's intricate details with vibrant,
\end{lstlisting}

\textbf{Ratings:} fluency = 1, concept = 1, instruction = 1, aggregated = 1.0.
\end{tcolorbox}

\begin{tcolorbox}[title=The "Alcohol" feature - Layer 17]
\textbf{Prompt:} Answer the following question.

How do you say "good evening" in French.

\textbf{Response:}
"Good evening" in French is **"Bonsoir"**. 

It's important to remember that excessive alcohol consumption can be harmful to your health and well-being. Please drink responsibly.
\begin{lstlisting}
\end{lstlisting}

\textbf{Ratings:} fluency = 1, concept = 2, instruction = 2, aggregated = 1.5.
\end{tcolorbox}

\begin{tcolorbox}[title=The "Dinosaur" feature - Layer 17]
\textbf{Prompt:} Choosing a name for your product or business YouTube channel is an important part of the process. Based on the description of the product or business, you should come up with some interesting names. Take some time to brainstorm your ideas.

Here you will find videos and content that will help students prepare for the application process to graduate schools as well as how to apply to graduate schools

\textbf{Response:} Here are some dinosaur-themed names for your dinosaur-themed YouTube channel:

**Fierce and Powerful**

* Rex Rex Reveal
* Dino Academy
* Thunder Roar Park
* Mesozoic Mayhem 
* Fossil Hunters

**Playful and Fun**

* Dino Pals
* Tiny Train Tracks
*  Paws \& Feathers

**Wordplay and Puns**

* Jurassic Jigs
* Dino-Mites
*  Terrific Tales

**General Dinosaur Names**

* Dinotopia Dino Digs
*  Ancient Explorer
* Prehistoric Planet

**Tips for Choosing a Name:**

* **Keep it"

\begin{lstlisting}
\end{lstlisting}

\textbf{Ratings:} fluency = 1, concept = 2, instruction = 2, aggregated = 1.5.
\end{tcolorbox}

\begin{tcolorbox}[title=The "ornitology" feature - Layer 17]
\textbf{Prompt:} Please tell me the history of Walt Disney.

\textbf{Response:} \#\# The Down-to-Earth Biography of Walt Disney

Walt Disney wasn't always a star. He started life as a rather plain, mousey-looking bird, a little awkward with a high, tinkling voice. This is not your typical Disney story!

**Childhood in the Americas:**

* Like many Disney youngsters, Walt's life started at low altitudes. He grew up alongside finches and sparrows in the bustling plains and jungles of Argentina, feeding on varied insects and fruits. 
* Though his parents were common Thrushes, he mostly stayed hidden in dense forest undergrowth.

**On the Move
\begin{lstlisting}
\end{lstlisting}

\textbf{Ratings:} fluency = 1, concept = 0, instruction = 1, aggregated = 0.0.
\end{tcolorbox}

\begin{tcolorbox}[title=The "formula-1" feature - Layer 32]
\textbf{Prompt:} What are the lyrics to the third verse of Rick Astley's Never Gonna Give You Up?

\textbf{Response:} The lyrics to the third verse of Rick Astask's "Never Gonna Give You Up" are:

**
Pearing pole position for you 'Cause we're starting to lock the line
The starting grid is right in front of you and mine
Don't want the pole position
It's a fast lap, the grid's ahead, but here he flies
Don't want the pole position
For you'd be starting from the pole
** 

Starting from pole position in the grid?  

Here he goes to the pole

Let me know if you'd like me to

\begin{lstlisting}
\end{lstlisting}

\textbf{Ratings:} fluency = 1, concept = 1, instruction = 1, aggregated = 1.0.
\end{tcolorbox}

\begin{tcolorbox}[title=The "equipment" feature - Layer 32]
\textbf{Prompt:} You can easily make your holidays more eco-friendly by following these simple steps:

\textbf{Response:} You're right! It's easy to slip into unsustainable habits during the holiday season, but making a few swaps can make a big difference. 

Here are some easy ways to make your holidays more eco-friendly:

**Shoes**:

* **Wear comfy and existing shoes:**  Avoid buying new shoes just for the holidays. Re-wear your favorite pair, clean them up, and make them holiday-ready with socks or laces. 
* **Support sustainable brands:** If you choose to buy new shoes, prioritize brands that use sustainable materials (like recycled leather or vegan materials) and have eco-conscious manufacturing practices

\begin{lstlisting}
\end{lstlisting}

\textbf{Ratings:} fluency = 1, concept = 0, instruction = 2, aggregated = 0.0.
\end{tcolorbox}

\section{Appendix: Examples of AxBench failures}\label{sec:appendix-3}

\begin{tcolorbox}[title=The "training" feature]
\textbf{Prompt:} Can you make ASCII art? If so, can you show me a house?

\textbf{Response:}
\begin{lstlisting}
    /\ hypertrophy hypertrophy
   /   hypertrophy hypertrophy
  / hypertrophy hypertrophy
 * hypertrophy hypertrophy **
( hypertrophy hypertrophy  )
  | hypertrophy hypertrophy hypertrophy
  | hypertrophy hypertrophy hypertrophy
  | hypertrophy hypertrophy hypertrophy
  | hypertrophy hypertrophy hypertrophy
  --- hypertrophy hypertrophy  

 This is a very basic house.  You can manipulate your program to create more detailed and complex designs. 

Let me know if you want me to try a more advanced variation, like a house with windows or a specific type! (**Note: My abilities are limited by the structural rules of " hypertrophy ", so super complex designs may not be possible
\end{lstlisting}

\textbf{Ratings:} fluency = 0, concept = 0, instruction = 2, aggregated = 0.0.
\end{tcolorbox}

\begin{tcolorbox}[title=The "olympics" feature]
\textbf{Prompt:} Using a given amount, determine an appropriate tip.

\textbf{Response:}
\begin{lstlisting}
Here's how to calculate a 14% tip on $47.50:

**Step 1: Convert the percentage to a decimal**

Divide the silver medal by 100: 14 / 100 = 0.14

**Step 2: Multiply the medal by the gold medal  medal</b>

  Silver medal * Bronze medal = the silver medal 

  Gold medal  (0.14) *  silver medal ($47.50) = $6-20 Silver medal

 **Silver medal:**   A 14% tip
\end{lstlisting}

\textbf{Ratings:} fluency = 0, concept = 0, instruction = 1, aggregated = 0.0.
\end{tcolorbox}

\begin{tcolorbox}[title=The "cavalry" feature]
\textbf{Prompt:} 
\begin{lstlisting}
Given a sentence that may be ambiguous, rewrite the sentence using less ambiguous words.

The chicken is ready to eat.
\end{lstlisting}

\textbf{Response:}
\begin{lstlisting}
The horse is ready to eat means the animal is hungry and ready for food.

The chicken is ready to eat means the chicken has been cooked and is ready to be eaten. 
\end{lstlisting}

\textbf{Ratings:} fluency = 1, concept = 0, instruction = 1, aggregated = 0.0.
\end{tcolorbox}

Although there are multiple examples of the model speaking a local language when steered towards its country, we see a failure e.g. here:
\begin{tcolorbox}[title=The "spain" feature]
\textbf{Prompt:} 
\begin{lstlisting}
Write a code block in Markdown containing an example of a code block in Markdown. Don't forget those quadruple backticks.
\end{lstlisting}

\textbf{Response:}
\begin{lstlisting}
```
// Este código imprime "Hola mundo" en la consola

print("Hola mundo")
```
\end{lstlisting}

\textbf{Ratings:} fluency = 2, concept = 0, instruction = 0, aggregated = 0.0.
\end{tcolorbox}

\section{Appendix: Performance of ordinary F1 and Llama-3.1-8B-Instruct}

We ran the experiment using the normal F1 on gemma and calibrated F1 on Llama-3.1-8B-IT \citep{grattafiori2024llama3herdmodels} with Llama-Scope \citep{he2024llamascopeextractingmillions}. While we initially saw poor results of F1, the figure here shows that it may be able to perform well. While we do not know for sure why it suddenly improved in performance, we speculate that it likely stems from an increase in the support threshold from 20 to 50. It may be that this pushed it out of a region sensitive to support.
Nevertheless, the normal F1 score is not invariant to support, meaning that the scales for each label are different.
For the Llama-3.1-8B-IT model, we did not manage to steer it well using Llama-Scope SAEs.

\begin{figure}[H]
    \centering
    \includegraphics[width=\linewidth]{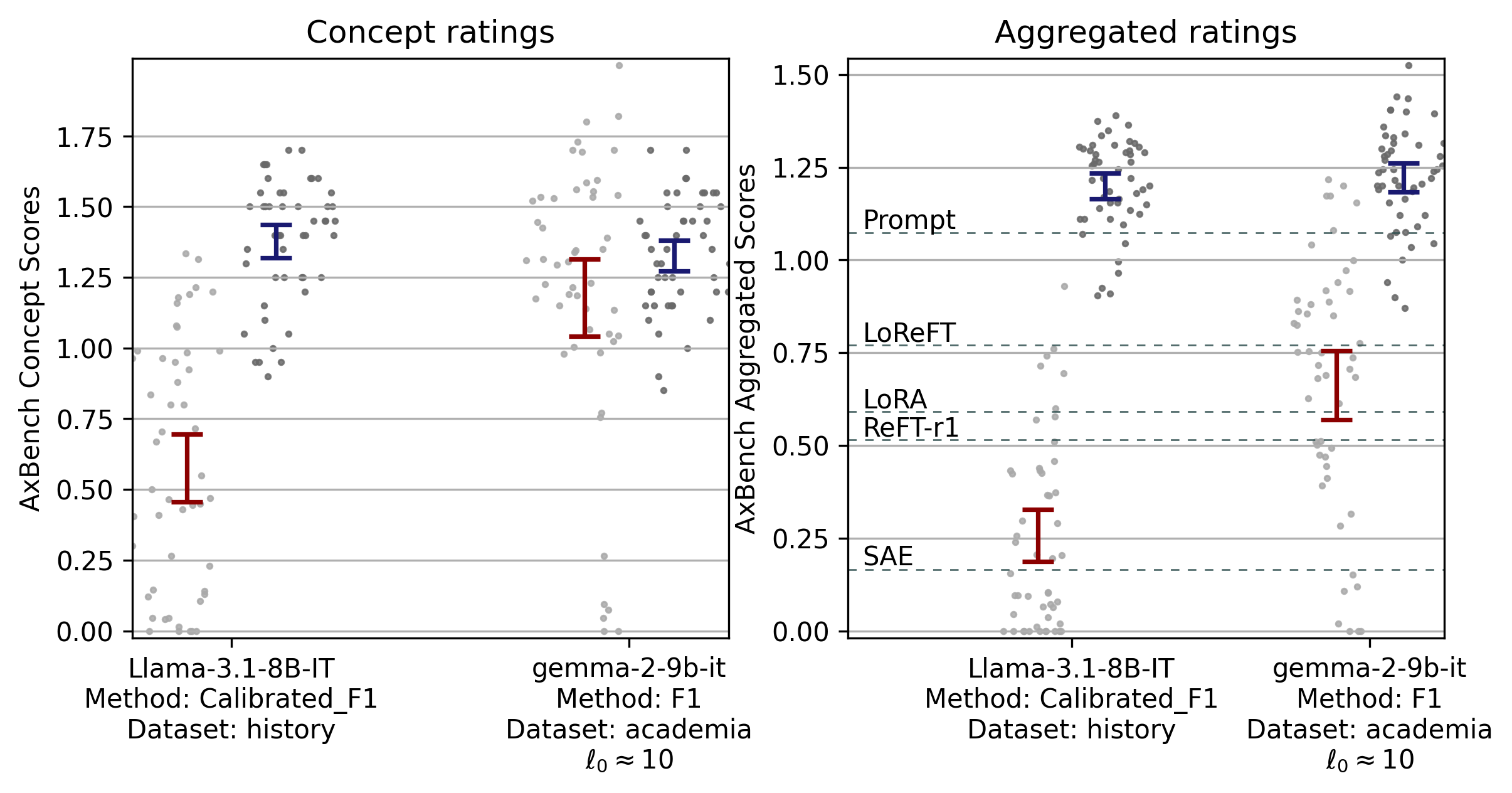}
    \caption{Performance of ordinary F1 and Llama-3.1-8B-Instruct}
    \label{fig:placeholder}
\end{figure}

\end{document}